%% file: PEF-7.tex
\documentclass[11pt]{article}
\usepackage{geometry}                
\usepackage{graphicx}
\usepackage{amsmath,amssymb, amsthm,epsfig,bm}
\usepackage{setspace}
\usepackage{epstopdf}
\usepackage{psfrag}
\usepackage{color,soul}
\usepackage{verbatim}
\usepackage{algorithm}
\usepackage{algpseudocode}
\usepackage{upgreek}
\usepackage{multirow}
\usepackage{natbib}
\usepackage{supertabular}
\usepackage{xr}

\usepackage{tikz}
\usetikzlibrary{arrows,shapes,trees, positioning}

\usepackage{hyperref}
\hypersetup{
    colorlinks=true,
    linkcolor=blue,
    filecolor=magenta,      
    urlcolor=purple,
    citecolor=blue,
}
\bibpunct{(}{)}{;}{a}{}{,} 
\sethlcolor{yellow}
\theoremstyle{plain} 
\newtheorem{theorem}{Theorem}

\newtheorem{proposition}[theorem]{Proposition}

\theoremstyle{definition}
\newtheorem{definition}{Definition}
\newtheorem{remark}{Remark}

\input{NewCommands}

\newcommand{\bdsZ}{\boldsymbol{Z}}

\begin{document}

\title{Learning Big Gaussian Bayesian Networks: \\
Partition, Estimation and Fusion
}
\author{Jiaying Gu and Qing Zhou\thanks{
Email: zhou@stat.ucla.edu;
this work was supported by NSF grant IIS-1546098.}\\
Department of Statistics, University of California, Los Angeles}
\date{}


\maketitle

\begin{abstract}

Structure learning of Bayesian networks has always been a challenging problem. Nowadays, massive-size networks with thousands or more of nodes but fewer samples frequently appear in many areas. We develop a divide-and-conquer framework, called partition-estimation-fusion (PEF), for structure learning of such big networks. The proposed method first partitions nodes into clusters, then learns a subgraph on each cluster of nodes, and finally fuses all learned subgraphs into one Bayesian network. The PEF method is designed in a flexible way so that any structure learning method may be used in the second step to learn a subgraph structure as either a DAG or a CPDAG. In the clustering step, we adapt the hierarchical clustering method to automatically choose a proper number of clusters. In the fusion step, we propose a novel hybrid method that sequentially add edges between subgraphs. Extensive numerical experiments demonstrate the competitive performance of our PEF method, in terms of both speed and accuracy compared to existing methods. Our method can improve the accuracy of structure learning by 20\% or more, while reducing running time up to two orders-of-magnitude.


\end{abstract}

\section{Introduction}\label{sec:PEFbackground}

The structure of a Bayesian network for $p$ random variables $X_1,\ldots,X_p$ is represented by a directed acyclic graph (DAG) $\mathcal{G}=(V,E)$. The node set $V=\left\{ 1,\ldots,p \right\}$ 
represents the set of random variables, and $E = \left\{(j,i) \in V \times V: j \rightarrow i\right\}$ is the edge set, where $j\rightarrow i$ is a directed edge in  $\mathcal{G}$. Let $\Pi_i^{\mathcal{G}} = \left\{j \in V: (j, i) \in E \right\}$	
denote the parent set of node $i$. The joint probability density function $f$ of
$\left( X_1,\ldots,X_p \right)$ can be factorized according to the structure of $\mathcal{G}$:
\begin{equation}\label{eq:BNdef}
f(x_1,\ldots,x_p)=\prod_{i=1}^{p}f(x_i|\pi_i),
\end{equation} 
where  $f(x_i|\pi_i)$ is the conditional probability density (CPD) of $X_i$ given $\Pi_i^{\mathcal{G}}=\pi_i$. 
Hereafter, we may use $X_i$ and the node $i$ interchangeably. 

The problem of structure learning of Bayesian networks from data has been an active research area due to its wide applications in machine learning, statistical modeling, and causal inference \citep{Spirtes93, Pearl00}. There are a few different approaches to this problem. The first one is the constraint-based approach, which determines the existence of edges by a sequence of conditional independence tests. 
The PC algorithm \citep{spirtes1991algorithm} and its further developments \citep{tsamardinos2003time,kalisch2007estimating, colombo2014order} are typical examples of constrained-based methods. 
The second category is so-called score-based learning, which searches for a graphical structure that optimizes a certain scoring function, such as early works in \cite{Heckerman95, geiger1994learning, chickering2002optimal, chickering2002finding}. Recently, fast algorithms have been developed to handle large and high-dimensional datasets \citep{Fu13, xiang2013lasso, aragam2015concave, ramsey2017million, zheng2018dags, Yuan19}. 
In addition, there are also hybrid methods that combine the above two approaches. These methods first restrict the search space using a constraint-based method, and then learn the DAG structure by optimizing a score over the restricted search space \citep{tsamardinos2006max,gamez2011learning,gasse2012experimental}.

Despite these great efforts, structure learning of Bayesian networks remains challenging, especially for datasets with a large number of variables. The DAG space grows super-exponentially in the number of nodes $p$ \citep{robinson1977counting}, and learning Bayesian networks has been shown to be an NP-hard problem in general \citep{chickering2004large}. Nowadays, it is common to generate and collect data from thousands of variables or more. As $p$ increases, however, many of the aforementioned methods slow down dramatically and become much less accurate, making them incompetent for large datasets. This motivates our development of a divide-and-conquer method that can learn massive-size Bayesian networks efficiently and accurately. Our method consists of three steps, Partition, Estimation and Fusion (PEF for short): 
\begin{enumerate}
\item \textit{P-step}: Partition the $p$ nodes into clusters based on a modified hierarchical clustering algorithm. 
\item \textit{E-step}: Apply an existing structure learning algorithm to estimate a subgraph on each cluster of nodes. 
\item \textit{F-step}: Develop a new hybrid method to merge the estimated subgraphs into a full DAG on all nodes. 
\end{enumerate}
Note that the number of nodes in a cluster is usually much smaller than $p$. This greatly speeds up structure learning in the estimation step, as most algorithms scale at least as $O(p^k)$ for some $k\geq 2$, e.g. \cite{kalisch2007estimating}. Moreover, this step can be parallelized in an obvious way, leading to further improvement in computational efficiency. The hybrid method in the fusion step first uses statistical tests to generate a candidate set of node pairs between estimated subgraphs, and then maximizes a modified BIC score by adding between-subgraph edges and updating within-subgraph edges. Since our conditional independence tests are performed based on the structure of subgraphs, the number of tests needed for our method is substantially smaller than a constraint-based method on a $p$-node problem. Our method is designed with maximum flexibility. The user can apply any structure learning algorithm in the second step as long as it outputs a PDAG (partially directed acyclic graphs), including DAGs and CPDAGs (completed PDAGs) as special cases. 

Our PEF method works very well on Bayesian networks with a block structure to some degree, having relatively weak connections between subgraphs. It is quite common for a large network to show such a block structure, due to the underlying heterogeneity among the nodes \citep{chin2015stochastic, decelle2011asymptotic, abbe2016exact}. From extensive numerical comparisons with existing methods, we find that the PEF method can significantly improve the accuracy of structure learning of Bayesian networks, while reducing computing time substantially, up to two orders-of-magnitude for big graphs.

The remaining of this paper is organized as follows. Section~\ref{sec:PEFprelim} contains a necessary background review for our method. Section~\ref{sec:PEFalgorithm} describes the partition and the estimation steps of the PEF method, while Section~\ref{sec:F-step} develops the fusion step in detail. Section~\ref{sec:PEFsimulation} provides numerical results of our method on real networks in comparison to other DAG learning algorithms. 
Section~\ref{sec:PEFdiscussion} summarizes this work with a discussion of future directions. Some technical details are deferred to an Appendix.

\section{Review of Bayesian networks}\label{sec:PEFprelim}

In this section, we briefly review some concepts about Bayesian networks that are most relevant to our method. The joint distribution $P$ that factorizes according to the DAG structure of a Bayesian network as in \eqref{eq:BNdef} satisfies so-called Markov properties \citep{lauritzen1996graphical}. Let $X,Y\in V$ and $\bdsZ\subseteq V\setminus\{X,Y\}$. If $\bdsZ$ $d$-separates $X$ from $Y$ in DAG $\mathcal{G}$, then the random variables $X$ and $Y$ are conditionally independent given $\bdsZ$. Using $\mathcal{D}_{\mathcal{G}}(X; Y|\boldsymbol{Z})$ to denote $d$-separation in $\calG$ and $\mathcal{I}_P(X; Y|\boldsymbol{Z})$ for conditional independence in $P$, the above (global) Markov property says that 
$\mathcal{D}_{\mathcal{G}}(X; Y|\boldsymbol{Z}) \Rightarrow \mathcal{I}_P(X; Y|\boldsymbol{Z})$.

\subsection{Faithfulness}

Note that the implication in a Markov property goes only in one direction. To estimate the structure of a DAG, we need to infer edges from conditional independence statements learned from data, which often requires the faithfulness assumption \citep{Spirtes93} to build up the equivalence between the two.

\begin{definition}[Faithfulness]\label{def:faithfulness}
Suppose $\mathcal{G}$ is a DAG equipped with a joint probability distribution $P$. Then $\mathcal{G}$ and $P$ are \textit{faithful} to each other if and only if 
\begin{equation*}
\mathcal{I}_P(X; Y|\boldsymbol{Z}) \Leftrightarrow \mathcal{D}_{\mathcal{G}}(X; Y|\boldsymbol{Z})
\end{equation*}
for any $X,Y\in V$ and $\bdsZ\subseteq V\setminus\{X,Y\}$. 
\end{definition}

If $(\mathcal{G}, P)$ satisfies the faithfulness assumption, we can use conditional independence (CI) test to infer $d$-separation in $\mathcal{G}$. Theorem~\ref{thm:CIstruct} provides a useful criterion to determine the existence of an edge using CI tests.

\begin{theorem}[\cite{Spirtes93}]\label{thm:CIstruct}
Suppose $(\mathcal{G}, P)$ satisfies the faithfulness assumption. Then there is no edge between a pair of nodes $X, Y \in V$ if and only if there exists a subset $\boldsymbol{Z} \subseteq V\setminus \{X,Y\}$ such that $\mathcal{I}_{P}(X; Y|\boldsymbol{Z})$.  
\end{theorem}

Consequently, faithfulness is commonly assumed in the development of many structure learning algorithms, especially constraint-based and hybrid methods, such as the PC algorithm and the MMHC algorithm \citep{Spirtes93,tsamardinos2006max}.

\subsection{Markov equivalence}\label{sec:obsEquivalence}

Multiple DAGs may imply the same set of $d$-separations, and thus encode the same set of CI statements, if they are \textit{Markov equivalent}:

\begin{definition}[Markov equivalence] \label{def:markovEqv}
Two DAGs $\mathcal{G}$ and $\mathcal{G}'$ on the same set of nodes $V$ are {Markov equivalent} if  
$\mathcal{D}_{\mathcal{G}}({X}; {Y}|\boldsymbol{Z}) \Leftrightarrow \mathcal{D}_{\mathcal{G}'}({X}; {Y}|\boldsymbol{Z})$ for any $X,Y\in V$ and $\bdsZ\subseteq V\setminus\{X,Y\}$.
\end{definition}


As shown by \cite{verma1990pearl}, two DAGs are Markov equivalent if and only if they have the same skeletons and the same $v$-structures. A $v$-structure is a triplet $\{i, j, k\} \subseteq V$ of the form $i \rightarrow k \leftarrow j$, where $i$ and $j$ are not adjacent, and the node $k$ is called an \textit{uncovered collider}. DAGs that are Markov equivalent form an equivalence class in the space of DAGs. A Markov equivalence class can be uniquely represented by a CPDAG \citep{chickering2002learning}.  The CPDAG for an equivalence class is defined as the PDAG consisting of directed edges for all compelled edges and undirected edges for all reversible edges in the equivalence class.

Since DAGs in the same equivalence class cannot be distinguished from observational data, some structure learning algorithms \citep{chickering2002finding, Spirtes93} output a CPDAG, instead of a particular DAG in the equivalence class. Thus, depending on which structure learning algorithm is used, the estimation step of our PEF method may output a DAG, a CPDAG, or in general a PDAG, from each cluster of nodes. The fusion step will merge these PDAGs into a full DAG as the final estimate.

\subsection{Gaussian Bayesian networks}

In this paper, we focus on Gaussian Bayesian networks for continuous data,
in which the conditional distributions are specified by a linear structural equation model,
\begin{eqnarray}\label{eq:defGaussModel}
X_j = \sum_{i\in \Pi_j^{\mathcal{G}}}\beta_{ij}X_i + \veps_j,\quad\quad j=1,\ldots,p,
\end{eqnarray}
where $\veps_j\sim \dnorm(0,\sigma_j^2)$ and $\beta_{ij} \neq 0$ if and only if $i \in \Pi_j^{\mathcal{G}}$. Let $B = (\beta_{ij})_{p\times p}$ be an edge coefficient matrix, which can be regarded as a weighted adjacency matrix for the DAG $\calG$, with $\beta_{ij}$ being the weight for the edge $i\to j$. Put $\Omg=\diag(\sigma_1^2,\ldots,\sigma_p^2)$ as a $p\times p$ diagonal matrix of error variances. Then the joint distribution of $(X_1,\ldots,X_p)$ defined by \eqref{eq:defGaussModel} is a multivariate Gaussian distribution $\dnorm_p(0,\Sigma)$ with covariance matrix
$\Sigma=(I-B)^{-\trans} \Omg (I-B)^{-1}$, where $I$ denotes the identity matrix. 

Suppose we have observed an iid sample of size $n$, $\mathbf{x}=[\mathbf{x}_{1}| \ldots | \mathbf{x}_{p}] \in{\R^{n \times p}}$, from a Gaussian Bayesian network parameterized by $(B,\Omg)$. Let $B_j$ be the $j$th column of $B$. Then the log-likelihood under this model is
\begin{equation}\label{eq:Gaussianll}
\ell(B, \Sigma) = \sum_{j=1}^p\left[-\frac{n}{2}\log(\sigma_j^2) - \frac{1}{2\sigma_j^2}\|\mathbf{x}_j - \mathbf{x}B_j\|^2\right],
\end{equation}
which forms the basis for score-based learning, subject to certain regularization or constraint on model complexity, e.g. the total number of edges in the DAG. For Gaussian random variables, conditional independence is equivalent to zero partial correlation, which is completely determined by the covariance matrix $\Sigma$. 
Consequently, in constraint-based methods, CI tests are performed based on sample partial correlations; see Appendix \ref{sec:partialCorr} for a brief description.

\section{Partition and estimation}\label{sec:PEFalgorithm}

In this section, we describe the first two steps of our PEF method. Besides some modifications to meet our specific needs in learning large networks, these two steps follow quite standard methods in clustering and structure learning. We will devote the entire Section~\ref{sec:F-step} to the fusion step.

\subsection{Partition}\label{sec:P-step}

As we have mentioned in the introduction, the first step (P-step) of our method is to partition nodes into clusters. Each node is associated with a data column $\bfx_j\in\R^n$ for $j=1,\ldots,p$. Let $C_i$, $i = 1, \ldots, k$, be the $k$ clusters generated by the P-step, and $S_i = |C_i|$ the size of the $i$th cluster. Accordingly, the underlying DAG $\mathcal{G}$ is cut into $k$ subgraphs. Let $s_w$ be the number of edges of $\calG$ within a subgraph, and $s_b$ the number of edges between subgraphs. In other words, $s_b$ is the number of edges in the partition-cut with respect to the $k$ clusters, which may be recovered later by the fusion step of our algorithm. In general, we want to control $s_b$ to a small value so that our recovery of the DAG structure will be more accurate. On the other hand, we wish that $k$ is quite large and the cluster size is as uniform as possible across the $k$ clusters, which will lead to maximum savings in computing time  for parallel learning of subgraphs in the E-step. 

To meet these specific needs for our problem, we propose a modified hierarchical clustering with average linkage that automatically chooses the number of clusters $k$. Define the distance between two nodes $i$ and $j$ by
\begin{equation} \label{eq:defDist}
d(i,j)= 1 - |r_{ij}| \in [0,1],
\end{equation}
where $r_{ij}=\cor(\bfx_i,\bfx_j)$ is the correlation between $\bfx_i$ and $\bfx_j$ for $i,j=1,\ldots,p$. 
\cite{hartigan1981consistency} suggests that one should only consider clusters with at least $5\%$ of the data points, which will be referred to as ``big clusters" hereafter. 

\begin{figure}[ht]
\begin{center}
\includegraphics[width=0.5\linewidth]{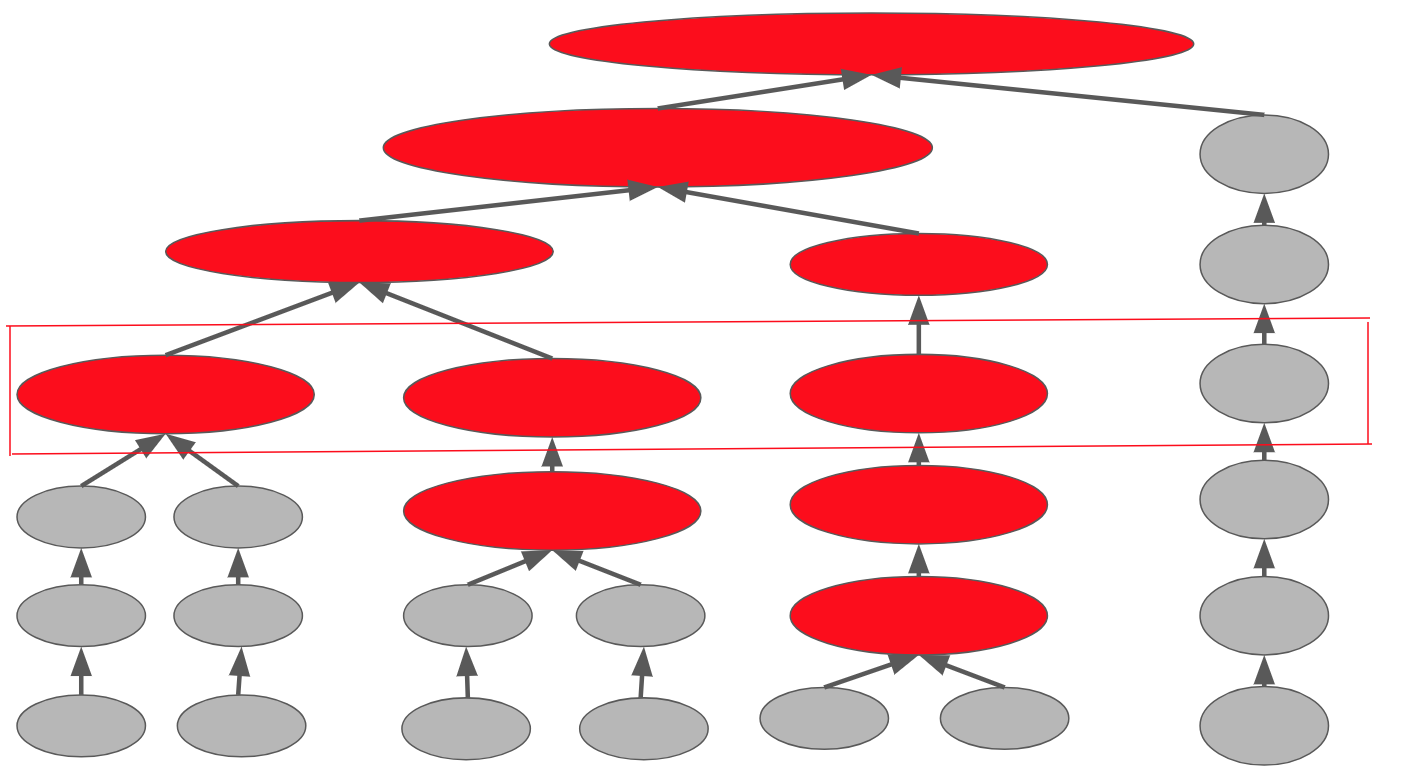}
\caption{Example for determining $k$ and $\ell$. Shown is the upper portion of a dendrogram. Red clusters are big clusters with more than $0.05p$ nodes, and the grey ones are small clusters. The level $\ell$ is marked by the red box, and in this case $k = 3$.}
\label{fig:cluster}
\end{center}
\end{figure}

Following this suggestion, we require the minimum cluster size be $0.05p$. As a result, there will be at most 20 clusters. Let $k_{\text{max}} \leq 20$ be the maximum number of clusters specified by the user. For $h = 0, 1, \ldots, p-1$, let $\mathcal{C}_h$ be the set of clusters formed at the $h$th step of the hierarchical clustering that proceeds in a bottom-up manner (Figure~\ref{fig:cluster}). In particular, $\mathcal{C}_0 = \{\{1\}, \{2\}, ..., \{p\}\}$ consists of $p$ singleton clusters and $\mathcal{C}_{p-1} = \{\{1, ..., p\}\}$ is just one cluster of all $p$ nodes. Let $k_i$ be the number of big clusters in $\mathcal{C}_i$. We choose
\begin{equation}\label{eq:chooseK}
k = \min\left\{k_{\max}, \underset{0\leq i\leq p-1}{\max} k_i \right\},
\end{equation}
which is the maximum number of big clusters subject to the user-specified $k_{\max}$.
Let $\ell$ be the highest level on the dendrogram with $k$ big clusters, i.e.
\begin{equation}\label{eq:chooseLevel}
\ell = {\argmax_{0\leq i\leq p-1}}\{i: k_i = k\}.
\end{equation}
Note that two big clusters will be merged at the next level ($\ell+1$) by the hierarchical clustering.
Figure~\ref{fig:cluster} shows an example of $k$ and $\ell$ on a dendrogram.

Relabel the clusters in $\mathcal{C}_{\ell}$ so that $S_1 \geq S_2 ... \geq S_{p-\ell}$, where $S_i = |C_i|$.
Then the first $k$ clusters are big clusters of interest. We assign the remaining small clusters to the $k$ big clusters by recursively merging two closest clusters if at least one of them is a small cluster. An outline of our clustering algorithm is shown in Algorithm~\ref{algo:cluster}. Note that in Line~\ref{line:assignCluster}, the distance $d(C_i, C_j) := \min\left\{d(X, Y): {X\in C_i, Y\in C_j}\right\}$, mainly for speed purpose. 

\begin{algorithm}[ht]
\caption{Modified hierarchical clustering}
\label{algo:cluster}
\begin{algorithmic}[1]
\State Hierarchical clustering given the dissimilarity matrix $D=(d(i,j))_{p\times p}$.
\State Generate the dendrogram $T_D$ of the hierarchical clustering.
\State Choose $k$ by \eqref{eq:chooseK} and $\ell$ by \eqref{eq:chooseLevel}.
\State Relabel clusters in $\mathcal{C}\leftarrow \mathcal{C}_\ell$ so that $S_1\geq ...\geq S_{p-\ell}$.
       \While{$|\mathcal{C}| > k$}
	\State $(i^*, j^*) \leftarrow \argmin_{(i, j)} \{d(C_i, C_j):i<j\;\text{and}\; j> k\}$. \label{line:assignCluster}
	\State $C_{i^*} \leftarrow C_{i^*}\cup C_{j^*}$, $\calC \leftarrow \calC\setminus \{C_{j^*}\}$.
       \EndWhile
\State Return $\mathcal{C} = \{C_1, C_2, ..., C_k\}$.
\end{algorithmic}
\end{algorithm}

\subsection{Estimation}\label{sec:E-step}

In the estimation step (E-step) we learn the structure of each subgraph individually. Under our PEF framework, this estimation step acts like a blackbox, and the user may use any structure learning algorithm to estimate the subgraphs without knowing its technical details. The output of this step is in general $k$ PDAGs. Note that both DAGs and CPDAGs are special cases of PDAGs.

In this work, we choose the CCDr algorithm \citep{aragam2015concave} in the \texttt{R} package \texttt{sparsebn} \citep{aragam2017Sparsebn} and the PC algorithm in the \texttt{R} package \texttt{pcalg} \citep{kalisch2012causal} as examples for the E-step. CCDr is a score-based method that outputs a DAG, while the PC algorithm is constraint-based and outputs a CPDAG or PDAG.  As such, we can illustrate the performance of the PEF method with different approaches, score-based versus constraint-based, to structure learning. We use the CCDr algorithm for two reasons: 1) It has competitive performance in terms of accuracy for structure learning of DAGs on high-dimensional data, which is our focus. 2) The way it is formulated and coded enables CCDr to learn quite large graphs, allowing for manageable comparisons with PEF in terms of running time. 

When the time complexity of a structure learning method grows faster than $O(p^2)$, the running time of learning small subgraphs in the E-step will be much shorter than estimating the full DAG as a whole. Furthermore, we can easily distribute the estimation step. Suppose in the partition step we have divided nodes into $k$ clusters $C_1,\ldots,C_k$, and the running time for learning a PDAG on $C_i$ is $t_i$. Learning $k$ subgraphs on $k$ cores in parallel will reduce the time for the E-step to $\max\{t_i, i=1,\ldots, k\}$, which is usually determined by the size of the largest cluster. According to our discussion in the previous subsection, 
there will be at most 20 clusters, and computing resource with 20 cores is very common nowadays. As supported by our numerical experiments, we can save majority of the computing time with the E-step.

\section{Fusion}\label{sec:F-step}

The fusion step (F-step) is a novel hybrid method developed to add edges between estimated subgraphs from the E-step and to learn the full DAG structure. It proceeds in two stages. First, we generate a candidate edge set $A$ to restrict our search space. By using a sequence of statistical tests, we identify a set $A^*$ of candidate edges between subgraphs. Then the candidate edge set $A$ consists of $A^*$ and all edges learned in each subgraph from the E-step. Second, we use a modified BIC score to learn the DAG structure by sequentially updating the edges in the set $A$. The final output of our PEF method is a DAG. 

\subsection{Candidate edge set}\label{sec:findA}


Recall that Theorem~\ref{thm:CIstruct} provides a justification for using conditional independence tests to infer edges of a DAG. In light of this result, we develop a method to produce a set $A^*$ of candidate edges between the subgraphs estimated from the E-step. Let $\mathcal{G}_m=(V_m,E_m)$, $m=1,\ldots,k$, denote these subgraphs and $z(i)\in\{1,\ldots,k\}$ the cluster label of node $i$.
In general, the subgraphs $\calG_m$ are PDAGs. We define the neighbors of a node $i$ in the subgraph $\mathcal{G}_{z(i)}$ as 
$$N_i(z(i)) = \{j\in V_{z(i)}:  j\rightarrow i \in E_{z(i)} \text{ or } (i, j) \in E_{z(i)} \},$$ 
where $j\rightarrow i$ denotes a directed edge and $(i, j)$ an undirected one. By Theorem~\ref{thm:CIstruct}, it is sufficient to find any subset of nodes $\bdsZ$ such that $X_i$ and $X_j$ are conditionally independent given $\bdsZ$ to conclude that there is no edge between $i$ and $j$. Unfortunately, for our problem size it is impractical to search all possible subsets.
To save calculation, we use the correlation $\tilde{\rho}_{ij} = \cor(\tilde{R}_i, \tilde{R}_j)$, where $\tilde{R}_i$ is the residual in projecting $X_i$ onto its neighbors $N_i(z(i))$ in $\mathcal{G}_{z(i)}$, to filter out unlikely between-subgraph edges. More specifically, we produce an initial candidate set 
\begin{align}\label{eq:tunePDAG}
\tilde{A}^* = \{(i, j): z(i)\ne z(j) \text{ and }\tilde{\rho}_{ij}=0 \text{ is rejected at significance level } \tilde{\alpha}\},
\end{align}
which will be refined further to define $A^*$. Proposition~\ref{thm:residual} shows that, under certain conditions, $\tilde{A}^*$ will include all between-subgraph edges if the test against $\tilde{\rho}_{ij}=0$ is perfect. Its proof can be found in Appendix \ref{sec:ProofPropo}.

\begin{proposition}\label{thm:residual}
Suppose the joint Gaussian distribution of $(X_1,\ldots,X_p)$ defined by \eqref{eq:defGaussModel} is faithful to the DAG $\calG$. Let $R_{i\cdot A}=X_i-\E(X_i\mid X_A)$ be the residual after regressing $X_i$ onto $X_A\defi(X_k)_{k\in A}$. If there is an edge $X_i\to X_j$ in $\calG$, then $R_{i\cdot A}$ and $R_{j\cdot B}$ are correlated for any disjoint $A,B\subseteq V\setminus \{i,j\}$ as long as $R_{i\cdot A}$ is independent of $X_A$ given $X_B$.
\end{proposition}

Since $R_{i\cdot A}$ is the residual after projecting $X_i$ onto $X_A$, by definition it is always independent of $X_A$. So the conclusion of the above proposition holds if $R_{i\cdot A}$ and $X_A$ do not become dependent after conditioning on $X_B$. 
Our rule \eqref{eq:tunePDAG} could produce false positive statements: $X_i$ and $X_j$ may become independent conditioning on other subsets. Therefore, we develop a sequential way to screen $\tilde{A}^*$ and define the final candidate edge set $A^*$ between subgraphs, described in Algorithm~\ref{algo:tune} Line~\ref{line:beginA} to Line~\ref{line:endA}: We go through each node pair $(i, j)\in\tilde{A}^*$ and run conditional independence test given the union of their updated neighbors, $N_i(z(i))\cup N_j(z(j))\cup P_{ij}$, where
\begin{equation}\label{eq:possibleP}
P_{ij} = \{k: (k, i)\in A^* \text{ or } (k, j)\in A^*\}
\end{equation}
is the set of neighbors of $i$ or $j$ in the current candidate set $A^*$ between subgraphs. 

\begin{algorithm}[!htb]
\caption{Find candidate edge set $A$}
\label{algo:tune}
\begin{algorithmic}[1]
\State Input data matrix $\mathbf{x}$ and estimated subgraphs $\mathcal{G}_1, ..., \mathcal{G}_k$. 
\State Set $\tilde{A}^* = \emptyset$. 
       \For {all pairs $(i,j)$ such that $z(i)\neq z(j)$} 
	    \If {$\tilde{\rho}_{ij}= 0$ is rejected at level $\tilde{\alpha}$} \label{line:testPhoTilde}
		\State $\tilde{A}^* \leftarrow \tilde{A}^* \cup (i,j)$.
                \EndIf
       \EndFor 
\State Set $A^* = \emptyset$.
	\For {all $(i, j) \in \tilde{A}^*$} \label{line:beginA}
		\State Let $\boldsymbol{Z} = N_i(z(i))\cup N_j(z(j))\cup P_{ij}$, where $P_{ij}$ is defined in \eqref{eq:possibleP}.
		\If {$\mathcal{I}_P(X_i; X_j|\boldsymbol{Z})$ is rejected at level $\alpha$} \label{line:condIndepTest}
			\State $A^* \leftarrow A^* \cup (i, j)$. \label{line:addtoA}
		\EndIf
	\EndFor \label{line:endA}
\State Return $A = A^* \cup SK(\mathcal{G})$. \label{line:attach}
\end{algorithmic}
\end{algorithm}

\begin{remark}\label{rem:sortActive}
In Algorithm~\ref{algo:tune}, node pairs are added to $A^*$ sequentially (Line~\ref{line:addtoA}) and thus, the result depends on the order we go through $\tilde{A}^*$. In our implementation, we sort the node pairs in $\tilde{A}^*$ in the ascending order of their $p$-values in testing against $\tilde{\rho}_{ij}= 0$ (Line~\ref{line:testPhoTilde}). In this way, node pairs that are more significant will have a higher priority to be included in the set $A^*$. Similarly, we also sort the node pairs in $A^*$ according to their $p$-values calculated in Line~\ref{line:condIndepTest}.
\end{remark}

Breaking a full DAG into subgraphs not only might introduce false negatives, i.e. the cut edges between two subgraphs, but also it could result in false positive edges within a subgraph. 
Suppose two non-adjacent nodes $i$ and $j$ share a common parent $k$ in the full DAG, but the P-step has put $k$ into a different cluster than $i$ and $j$. Then in the subgraph containing $i$ and $j$, there will be an edge $(i,j)$ in the estimated skeleton since they are {\em not} independent without conditioning on $k$. By cutting some of the edges in the P-step, we have changed the structure of a subgraph, and therefore a structure learning algorithm in the E-step may not recover the true subgraph in the original DAG. To fix this problem, we will revisit all edges learned from the E-step and correct the subgraph structures based on the new edges added between subgraphs. Let $\mathcal{G}=(V,E)$ be the PDAG consisting of disconnected subgraphs learned from the E-step and $SK(\mathcal{G})$ be the (undirected) edge set in the skeleton of $\mathcal{G}$, i.e., 
$$SK(\mathcal{G})=\{(i,j): (i,j) \in E \text{ or }i\to j \in E\}.$$ 
Our candidate edge set $A$ is formed by attaching $SK(\mathcal{G})$ to the end of $A^*$ (Line~\ref{line:attach} in Algorithm~\ref{algo:tune}). The edges of our final output DAG will be restricted to a subset of $A$. The complete algorithm for finding the candidate edge set $A$ is summarized in Algorithm~\ref{algo:tune}.

\subsection{Learning full DAG structure} \label{sec:highDimBIC}

The last stage in the fusion step is to determine, for each node pair $(i,j)\in A$, whether there is an edge and its orientation if an edge does exist. This is done by minimizing a modified BIC score, called the risk inflation criterion (RIC) \citep{foster1994risk}, over the candidate edge set in a sequential manner. The RIC score has two components,  a log-likelihood part to measure how good a graph $\calG$ fits the data and a regularization term to enforce sparsity:
\begin{equation}\label{eq:BIC}
\mathrm{RIC}(\calG) = -2\ell(\widehat{B},\widehat{\Omg}\mid \mathcal{G}) + \lambda d(\mathcal{G}),
\end{equation}
where $\ell(\cdot\mid \mathcal{G})$ is the log-likelihood \eqref{eq:Gaussianll} evaluated at the MLE $(\widehat{B},\widehat{\Omg})$ given the DAG $\calG$, $d(\mathcal{G})$ is the number of edges, and 
$\lambda = 2\log p$. We use this score when the number of nodes is large with $p>\sqrt{n}$. When $p\leq\sqrt{n}$, we switch back to the regular BIC score, i.e. $\lambda=\log n$. 

For each $(i, j)\in A$, we need to compare three models:
\begin{eqnarray}\label{eq:modelCompare}
& & M_0:  \text{no edge between } i \text{ and } j, \nonumber\\
& & M_1:  i \text{ is a parent of } j, \\
& & M_2:  j \text{ is a parent of } i, \nonumber
\end{eqnarray}
while holding other edges in $\calG$ fixed. Since the RIC score \eqref{eq:BIC} is decomposable, this comparison reduces to comparing the score difference for the involved child nodes (see Appendix \ref{sec:RIC}). If there is a true edge between the two nodes $i$ and $j$, both $M_1$ and $M_2$ will have a lower RIC score than $M_0$ in the large sample limit, due to the nonzero partial correlation between $X_i$ and $X_j$ given any other set of variables. Thus, we will add an edge between $(i, j)$ if and only if
\begin{equation}\label{eq:BICcriterion}
\max\left\{\mathrm{RIC}(M_1), \mathrm{RIC}(M_2)\right\} < \mathrm{RIC}(M_0),
\end{equation}
where $\mathrm{RIC}(M)$ is the RIC score for model $M$. If criterion \eqref{eq:BICcriterion} is met, we will further decide the edge orientation. To enforce acyclicity, if the edge $i\to j$ (or $j\to i$) induces a directed cycle, we add $j\rightarrow i$ (or $i\to j$). If neither direction induces a directed cycle, we choose the model with a smaller RIC following a default tie-breaking rule. See Appendix \ref{sec:RIC} for more technical details. 

The full fusion step is shown in Algorithm~\ref{algo:fusion}, which cycles through $A$ iteratively until the structure of $\mathcal{G}$ does not change. Denote by $N_i(\cal{G})$ the neighbors of node $i$ in the current $\mathcal{G}$. At any iteration, if $\mathcal{I}_P(X_i; X_j|N_i(\mathcal{G})\cup N_j(\mathcal{G}))$ according to the conditional independence test (Line~\ref{line:secondTune}), we will remove the pair $(i,j)$ from $A$ permanently. This rule is again justified by Theorem~\ref{thm:CIstruct} under faithfulness. In order to reduce the number of false positive edges for large $p$, the significance level for all tests, including $\tilde{\alpha}$ and $\alpha$ in Algorithm~\ref{algo:tune}, is set to $0.001$ in our implementation.

\begin{algorithm}[!htb]
\caption{Fuse subgraphs}
\label{algo:fusion}
\begin{algorithmic}[1]
\State Input data matrix $\mathbf{x}$ and estimated subgraphs $\mathcal{G}_1, ..., \mathcal{G}_k$. 
\State Run Algorithm~\ref{algo:tune} to generate candidate edge set $A$.
\State Initialize $\mathcal{G}$ to be the PDAG consisting of $\calG_1,\ldots,\mathcal{G}_k$.
	\For {all $(i, j)\in A$} \label{line:beginBIC}
		\If {$i, j$ are adjacent in $\mathcal{G}$}
			\State remove the edge from $\mathcal{G}$.
		\EndIf
		\If {$\mathcal{I}_P(X_i; X_j|N_i(\mathcal{G})\cup N_j(\mathcal{G}))$} \label{line:secondTune} 
			\State $A\leftarrow A\setminus \{(i,j)\}$. 
		\Else
			\State $\mathrm{RIC}_{\max} = \max\left(\mathrm{RIC}(M_1), \mathrm{RIC}(M_2)\right)$.
			\If {$\mathrm{RIC}_{\max} < \mathrm{RIC}(M_0)$}
				\If{adding edge $i\rightarrow j$ induces a cycle}
					\State add $j \rightarrow i$ to $\mathcal{G}$
				\ElsIf {adding edge $j\rightarrow i$ induces a cycle}
                  				 \State add $i \rightarrow j$ to $\mathcal{G}$
				\Else 
					\State choose the direction that leads to a smaller RIC.
				\EndIf
			\EndIf
		\EndIf
	\EndFor \label{line:endBIC}
\State Repeat \ref{line:beginBIC} to \ref{line:endBIC} until the structure of $\mathcal{G}$ does not change and return $\mathcal{G}$.
\end{algorithmic}
\end{algorithm}

\section{Numerical experiments}\label{sec:PEFsimulation}

In this section, we test our PEF method on Gaussian data generated from real networks. We choose to use two different structure learning algorithms in the E-step, the CCDr algorithm \citep{aragam2015concave} which estimates a DAG and the PC algorithm which outputs a PDAG. We will call these two implementations PEF-CCDr and PEF-PC hereafter. Accordingly, we compare the results from the PEF methods with those from the CCDr and the PC algorithms applied on the whole data, which will demonstrate the advantages of our divide-and-conquer strategy in learning large networks.

\subsection{Data generation} \label{sec:datGen}

All network structures were downloaded from the repository of the \texttt{R} package \texttt{bnlearn} \citep{scutari2010learning, Scutari2017learnining}. The networks used in this work are: PATHFINDER, ANDES, DIABETES, PIGS, LINK, and MUNIN, with the number of nodes $p = (135, 223, 413, 441, 724, 1041)$ and the number of edges $s_{\text{sub}} = (195, 338, 602, 592, 1125, 1397)$. In order to generate large DAGs, we replicate each network $k$ times and randomly add some edges between copies of the network. For easy reference, define $\text{Net}(k, c)$ to be the DAG composed of $k$ replicates of Net with $c\cdot ks_{\text{sub}}$ edges added between subgraphs, where $c \geq 0$ is a constant and Net is one of the above six networks. Let $s_w = ks_{\text{sub}}$ be the number of within-subgraph edges and $s_b$ be the number of between-subgraph edges. Then $c=s_b/s_w$ is the ratio between the numbers of the two types of edges. For example, ANDES$(5,0.1)$ refers to a network constructed by 5 copies of the ANDES network with $s_b=0.1s_w$ edges added between the 5 sub-networks.  
Denote the number of true edges in a DAG by $s_0 = s_w+s_b$. We have three network generation schemes:
\begin{itemize}
\item[  i.] Net$(5,c)$ for $c \in\{ 0, 0.1\}$ and Net$\in\{$PATHFINDER, ANDES, DIABETES, PIGS, LINK$\}$. 
In total, ten networks were generated by this scheme. 
\item[ ii.] Mixed networks: We combined networks PATHFINDER, ANDES, DIABETES, PIGS, LINK to build a DAG with $k=5$ different subgraphs. Similar to scheme (i), we randomly added $s_b =c s_w$ edges between subgraphs for $c \in\{0, 0.1\}$. We refer to these two networks as Mix$(5, c)$.
\item[iii.] MUNIN$(k,0)$ for $k = 1, \ldots, 10$: MUNIN is the largest network available on the \texttt{bnlearn} repository. We did not add any edges between the subgraphs. So the number of edges for each DAG generated here was $s_0 = s_w = ks_{\text{sub}}$. 
\end{itemize}


Data sets from the above DAGs were generated according to the linear structural equation model in \eqref{eq:defGaussModel}. We drew $\beta_{ij}$ uniformly from $[-1, -0.5]\cup [0.5, 1]$ if $(i, j)\in E$ and set $\beta_{ij} = 0$ otherwise. The error variances $\sigma_j^2, j=1, ..., p$ were chosen so that all data columns had the same standard deviation. The number of observations for all simulated data sets were set to $n=1,000$. For each network generated by schemes (i) and (ii), we simulated 10 data sets. Networks in scheme (iii) were mainly used to test the limit of structure learning algorithms, so only 5 data sets were generated from each DAG.

\subsection{Accuracy metrics}

We propose a few metrics to evaluate the accuracy of PDAGs learned by structure learning algorithms. As DAGs can be regarded as a special class of PDAGs, metrics defined here can be used to assess the quality of estimated DAGs as well. Since we are using observational data, structure learning algorithms may not determine all edge orientations due to Markov equivalence (Definition~\ref{def:markovEqv}). In our assessment, we take $v$-structures and compelled edges into account in the following definitions of accuracy metrics:
\begin{itemize}
\item[-] T, the number of edges in the true graph.
\item[-] P, the number of predicted edges by a structure learning algorithm.
\item[-] E, the number of expected edges for which the true graph and the estimated graph coincide. We define an estimated directed edge to be expected if it meets either of the following two criteria: (1) This edge is in the true DAG with the correct orientation; (2) The edge coincides after converting the estimated DAG and the true DAG to CPDAGs.  An estimated undirected edge is considered expected if it satisfies condition 2. 
\item[-] R, the number of reversed edges. This is the number of predicted edges in the true skeleton, excluding expected edges.
\item[-] FP $=$ P $-$ E $-$ R, the number of false positive edges.
\item[-] SHD $=$ R $+$ M $+$ FP, the structural Hamming distance between the estimated and the true graphs, where M $=$ T $-$ E $-$ R is the number of missing edges.
\item[-] JI $=$ E$/($T $+$ P $-$ E$)$, the Jaccard index, i.e. the ratio of the number of common edges over the size of the union of the edge sets of two graphs.
\end{itemize}
In particular, SHD and JI are overall accuracy metrics. Small SHD and high JI indicate high accuracy in structure learning.

\subsection{Comparison with the CCDr algorithm}\label{sec:PEFccdrSim}

We will first show the improvement in speed of our PEF-CCDr method compared to the CCDr algorithm. Then we will show that the PEF-CCDr method actually improves the accuracy of the CCDr algorithm. For all the experiments, we ran  CCDr provided in the \texttt{R} package \texttt{sparsebn}. The CCDr algorithm outputs a solution path with an increasing number of edges. In order to enforce sparsity, we simply chose the DAG along the solution path with around $1.5p$ edges, and stopped running CCDr when the number of estimated edges on the path became greater than $2p$ by setting \texttt{edge.threshold} $= 2p$. Note that we used exactly the same settings for  CCDr applied to learn the full graph and in the E-step of the PEF method. 

\subsubsection{Timing comparison}

\begin{figure}[t]
\begin{minipage}[c]{0.5\textwidth}
\begin{center}
\includegraphics[scale=0.4]{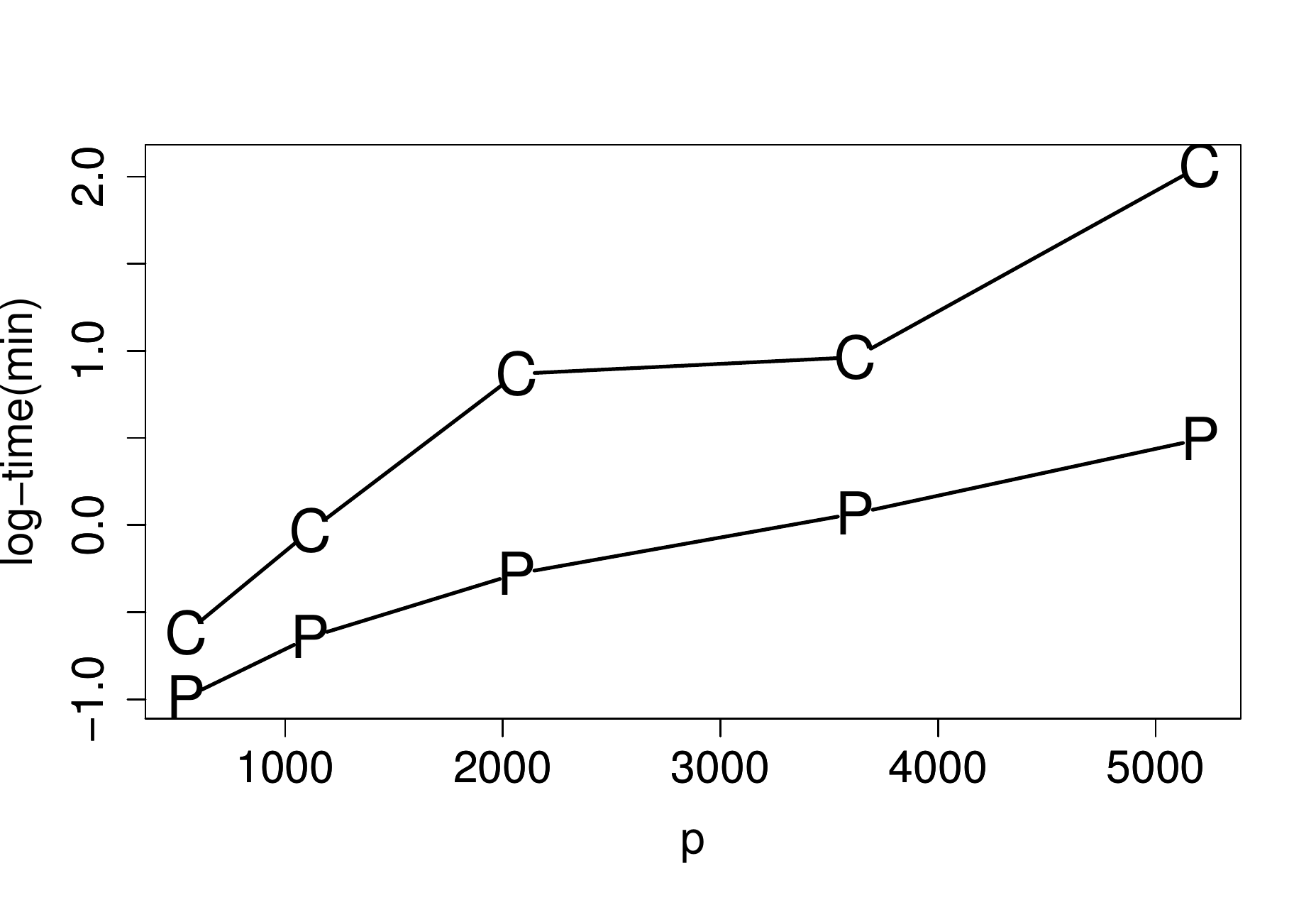} \\
{(a)}
\end{center}
\end{minipage}\hfill
\begin{minipage}[c]{0.5\textwidth} 
\begin{center}
\includegraphics[scale=0.4]{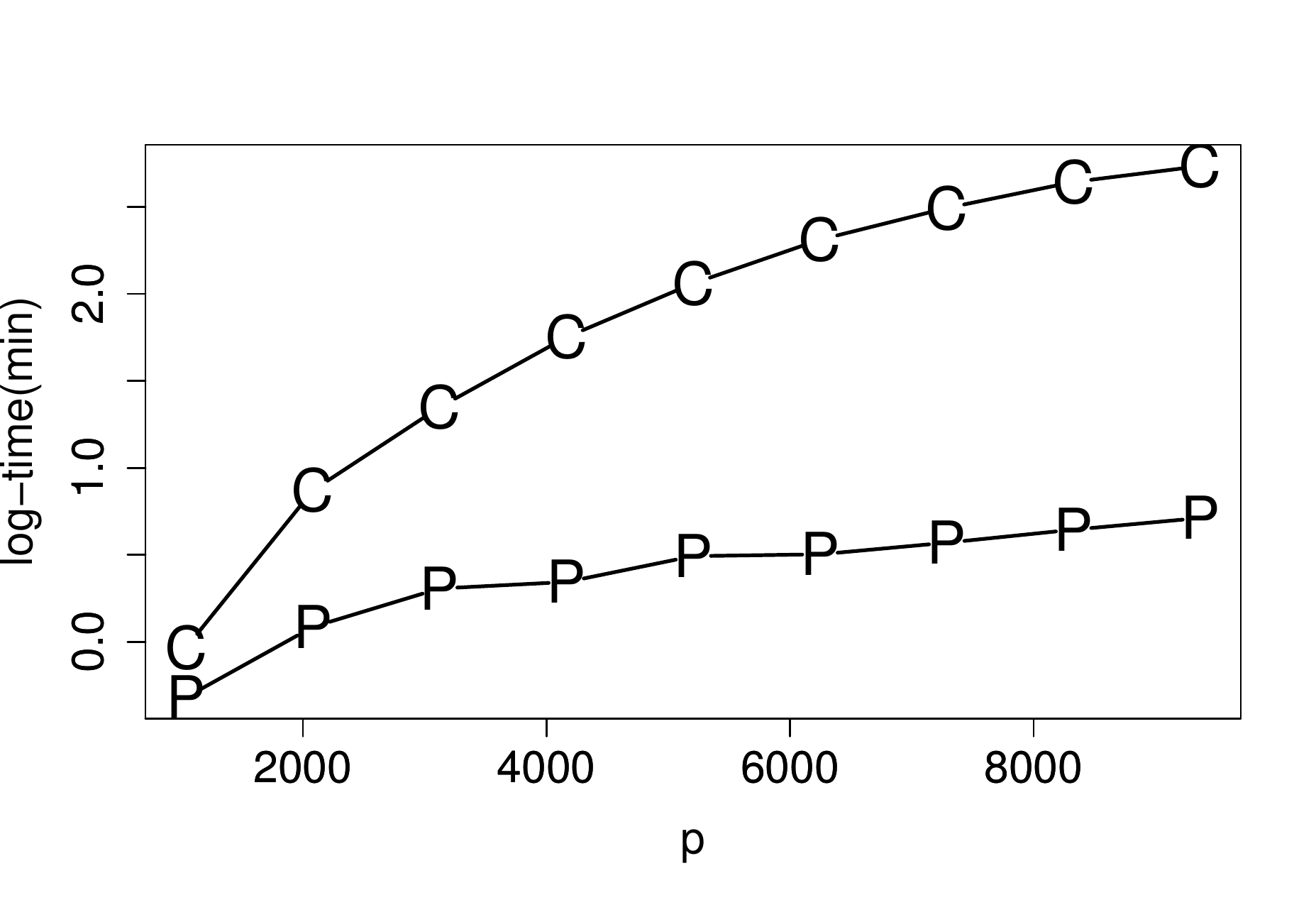} \\
{(b)}
\end{center}
\end{minipage}
\caption{Log$_{10}$ running time for different size of DAGs. The line with -C- is for CCDr and the line with -P- for PEF-CCDr.}
\label{fig:timeCmp}
\end{figure}

Figure~\ref{fig:timeCmp} reports the $\log_{10}$ running times of the two algorithms. Figure~\ref{fig:timeCmp}(a) illustrates how the two methods scaled when the size of the subgraphs increased, tested on the networks Net$(5, 0)$ for Net $\in \{$PATHFINDER, ANDES, DIABETES, LINK, MUNIN$\}$. Figure~\ref{fig:timeCmp}(b) illustrates how the two methods scaled when the number of subgraphs increased, using the networks MUNIN$(k, 0)$ for $k = 1, ..., 10$. Table~\ref{tab:timePefCcdr} reports the total running time (T) of  CCDr and  PEF-CCDr, as well as the running time of each step  (P, E, F) of PEF-CCDr for all 22 networks (Section~\ref{sec:datGen}). For the E-step in our PEF-CCDr method, we report the time for parallel estimation of multiple subgraphs. 

\begin{table}[!h]
\begin{center}
\caption{Timing comparison (in minutes) between CCDr and PEF-CCDr}
\label{tab:timePefCcdr}       
\begin{tabular}{lr | r | rrrr | r | r}
\noalign{\smallskip}\hline\noalign{\smallskip}
 & & CCDr  & \multicolumn{5}{c|}{PEF-CCDr} &  \\
\noalign{\smallskip}\cline{3-9}\noalign{\smallskip}
Network & $p$ & T & T & P & E & F & $\hat{k}$ & $r_T$ \\
\noalign{\smallskip}\hline\noalign{\smallskip}
PATHFINDER(5, 0) & 545 & 0.24 & 0.10 & 0.01 & 0.01 & 0.08 & 5.0 & 2.40 \\
PATHFINDER(5, 0.1) & 545 & 0.23 & 0.10 & 0.01 & 0.01 & 0.08 & 5.0 & 2.30 \\
ANDES(5, 0) & 1115 & 0.93 & 0.24 & 0.02 & 0.02 & 0.20 & 9.5 & 3.88 \\
ANDES(5, 0.1) & 1115 & 0.59 & 0.38 & 0.02 & 0.02 & 0.34 & 8.4 & 1.55 \\
MIX(5, 0) & 1910 & 4.65 & 0.46 & 0.06 & 0.08 & 0.32 & 8.6 & 10.11 \\
MIX(5, 0.1) & 1910 & 2.51 & 0.67 & 0.06 & 0.16 & 0.45 & 7.2 & 3.75 \\
DIABETES(5, 0) & 2065 & 7.38 & 0.53 & 0.06 & 0.09 & 0.38 & 8.1 & 13.92 \\
DIABETES(5, 0.1) & 2065 & 4.73 & 0.60 & 0.05 & 0.08 & 0.47 & 8.2 & 7.88 \\
PIGS(5, 0) & 2205 & 10.84 & 0.64 & 0.07 & 0.16 & 0.41 & 5.8 & 16.94 \\
PIGS(5, 0.1) & 2205 & 6.60 & 0.74 & 0.07 & 0.14 & 0.53 & 6.2 & 8.92 \\
LINK(5, 0) & 3620 & 9.19 & 1.17 & 0.17 & 0.16 & 0.84 & 8.1 & 7.85 \\
LINK(5, 0.1) & 3620 & 9.90 & 1.59 & 0.16 & 0.17 & 1.26 & 9.2 & 6.23 \\
\noalign{\smallskip}\hline\noalign{\smallskip}
MUNIN(1, 0) & 1041 & 0.93 & 0.48 & 0.02 & 0.20 & 0.27 & 7.0 & 1.94 \\
MUNIN(2, 0) & 2082 & 7.42 & 1.22 & 0.07 & 0.79 & 0.36 & 4.2 & 6.08 \\
MUNIN(3, 0) & 3123 & 22.32 & 2.03 & 0.12 & 1.04 & 0.87 & 3.0 & 11.00 \\
MUNIN(4, 0) & 4164 & 56.42 & 2.21 & 0.22 & 1.13 & 0.86 & 4.0 & 25.53 \\
MUNIN(5, 0) & 5205 & 114.85 & 3.11 & 0.34 & 1.21 & 1.57 & 5.0 & 36.93 \\
MUNIN(6, 0) & 6246 & 204.93 & 3.18 & 0.46 & 1.28 & 1.44 & 6.0 & 64.44 \\
MUNIN(7, 0) & 7287 & 311.59 & 3.71 & 0.64 & 1.28 & 1.79 & 7.0 & 83.99 \\
MUNIN(8, 0) & 8328 & 440.02 & 4.42 & 0.82 & 1.26 & 2.33 & 8.0 & 99.55 \\
MUNIN(9, 0) & 9369 & 542.56 & 5.15 & 1.04 & 1.33 & 2.78 &9.0 & 105.35 \\
MUNIN(10, 0) & 10410 & NA & 5.94 & 1.32 & 1.39 & 3.23 & 10.0 & NA \\
\noalign{\smallskip}\hline
\end{tabular}\\
\end{center}
{Note: $p$ is the number of nodes, T is the total running time, P, E, and F are the running times for the P-step,  the E-step, and the F-step, respectively. $\hat{k}$ is the average number of estimated clusters in the P-step. $r_T$ is the ratio of total running time of  CCDr over that of PEF-CCDr.}
\end{table}

From Figure~\ref{fig:timeCmp}(a) we see that when the number of subgraphs stayed the same and the size of the sub-graphs became larger, the running time of PEF-CCDr increased monotonically. The scalability of the E-step depends on the CCDr algorithm. Therefore, the running time of the E-step of our PEF-CCDr method increased with the size of the subgraphs, in a similar pattern as the CCDr algorithm did. As reported in Table~\ref{tab:timePefCcdr}, for the largest network MUNIN$(5, 0)$ included in Figure~\ref{fig:timeCmp}(a), PEF-CCDr was 37 times faster than CCDr. 

From the lower panel of Table~\ref{tab:timePefCcdr} as well as Figure~\ref{fig:timeCmp}(b), we see that as $k$ increased, improvement of our PEF method in speed became more substantial. The number of clusters our PEF-CCDr method identified ($\hat{k}$) is shown in Table~\ref{tab:timePefCcdr}. The PEF-CCDr method identified the correct number of subgraphs for $k \geq 3$, and therefore the running time of the E-step stayed comparable to the running time of CCDr on a single MUNIN network (around 1 minute). When the number of subgraphs $k= 3$, PEF-CCDr was 11 times faster than CCDr, and when $k=9$, it was 105 times faster. When $k$ was increased to 10, our device for running the tests, MacBook Pro with 3.1 GHz Intel Core i7 processor, ran out of memory for the CCDr algorithm. Our PEF-CCDr method, on the other hand, took only 5.94 minutes to run MUNIN(10, 0). 
This example shows the huge advantage of our PEF-CCDr method in terms of computational efficiency for learning big Bayesian networks. 

\subsubsection{Accuracy comparison}\label{sec:cmpCCDr}

Next, we compare the accuracy between the PEF-CCDr method and the CCDr algorithm.
Table~\ref{tab:pathfinderAndes} reports the summary of accuracy for the two methods on ten networks generated in the first two schemes (Section~\ref{sec:datGen}). In this and subsequent tables, Net$(5)$ refers to either Net$(5,0)$ or Net$(5,0.1)$ with the value of $c$ implicitly given by $(s_0,s_b)$.
We see from Table~\ref{tab:pathfinderAndes} that for all cases the SHD of PEF-CCDr was much smaller than CCDr, and the JI was higher than CCDr.
For PATHFINDER(5) with $p = 545<n=1000$, the advantage of our PEF-CCDr method was not as obvious as the rest of the big networks.
For all other networks where $p > n$, the number of expected edges of our PEF-CCDr method increased  more than $15\%$ compared to CCDr in most of the cases, while the reversed edges and the false positives decreased more than $20\%$. The overall metrics SHD decreased more than $20\%$ and the JI increased over $35\%$ for all cases. 

\begin{table*}[!htb]
\begin{center}
\caption{Accuracy comparison between CCDr and PEF-CCDr}
\label{tab:pathfinderAndes}       
{\renewcommand{\arraystretch}{0.95}
\begin{tabular}{llrrrrrr}
\noalign{\smallskip}\hline\noalign{\smallskip}
$(s_0, s_b)$ & Method & P & E & R & FP &  SHD & JI \\
\noalign{\smallskip}\hline\noalign{\smallskip}
\multicolumn{8}{c}{PATHFINDER$(5)$, $p = 545$} \\
\noalign{\smallskip}
{$(975, 0)$} & CCDr & 823.0 & 252.7 & 149.6 & 420.7 & 1143.0 & 0.164 \\
& PEF-CCDr & 660.4 & 254.7 & 122.4 & 283.3 & 1003.6 & 0.186 \\
{$(1073, 98)$} & CCDr & 838.0 & 329.5 & 123.4 & 385.1 & 1128.6 & 0.209  \\
& PEF-CCDr & 768.5 & 361.2 & 119.1 & 288.2 & 1000.0 & 0.245 \\

\noalign{\smallskip}
\multicolumn{8}{c}{ANDES$(5)$, $p = 1115$} \\
\noalign{\smallskip}
{$(1690, 0)$} & CCDr & 1586.0 & 931.4 & 447.0 & 207.6 & 966.2 & 0.397 \\
& PEF-CCDr & 1563.0 & 1187.7 & 224.3 & 151.0 & 653.3 & 0.576 \\
{$(1859, 169)$} & CCDr & 1721.8 & 1051.6 & 452.1 & 218.1 & 1025.5 & 0.416 \\
& PEF-CCDr & 1766.1 & 1406.1 & 186.6 & 173.4 & 626.3 & 0.634 \\

\noalign{\smallskip}
\multicolumn{8}{c}{DIABETES$(5)$, $p = 2065$} \\
\noalign{\smallskip}
{$(3010, 0)$} & CCDr & 3166.3 & 1327.3 & 1067.9 & 771.1 & 2453.8 & 0.274 \\
& PEF-CCDr & 2779.8 & 1580.2 & 779.1 & 420.5 & 1850.3 & 0.376 \\
{$(3311, 301)$} & CCDr & 3069.6 & 1499.4 & 978.9 & 591.3 & 2402.9 & 0.307 \\
& PEF-CCDr & 3202.7 & 2010.1 & 702.4 & 490.2 & 1791.1 & 0.447 \\

\noalign{\smallskip}
\multicolumn{8}{c}{PIGS$(5)$, $p = 2205$} \\
\noalign{\smallskip}
{$(2960, 0)$} & CCDr & 3285.6 & 1677.4 & 832.0 & 776.2 & 2058.8 & 0.367 \\
& PEF-CCDr & 2809.9 & 1933.5 & 541.6 & 334.8 & 1361.3 & 0.504 \\
{$(3256, 296)$} & CCDr & 3262.5 & 1874.0 & 800.8 & 587.7 & 1969.7 & 0.404 \\
& PEF-CCDr & 3182.1 & 2308.3 & 489.9 & 383.9 & 1331.6 & 0.559 \\

\noalign{\smallskip}
\multicolumn{8}{c}{LINK$(5)$, $p = 3620$} \\
\noalign{\smallskip}
{$(5625, 0)$} & CCDr & 5329.4 & 2640.6 & 1421.7 & 1267.1 & 4251.5 & 0.318 \\
& PEF-CCDr & 5021.9 & 3211.4 & 972.6 & 837.9 & 3251.5 & 0.432 \\
{$(6188, 563)$} & CCDr & 5799.6 & 3096.9 & 1436.5 & 1266.2 & 4357.3 & 0.348 \\
& PEF-CCDr & 5849.9 & 4018.3 & 878.1 & 953.5 & 3123.2 & 0.501 \\

\noalign{\smallskip}
\multicolumn{8}{c}{Mix$(5)$, $p = 1910$} \\
\noalign{\smallskip}
{$(2852, 0)$} & CCDr & 2893.1 & 1423.4 & 766.4 & 703.3 & 2131.9 & 0.329 \\
& PEF-CCDr & 2620.2 & 1685.9 & 526.2 & 408.1 & 1574.2 & 0.446 \\
{$(3138, 286)$} & CCDr & 2923.5 & 1564.6 & 766.0 & 592.9 & 2166.3 & 0.348 \\
& PEF-CCDr & 2965.3 & 2005.0 & 497.5 & 462.8 & 1595.8 & 0.489 \\
\noalign{\smallskip}\hline
\end{tabular}%
}
\end{center}
\end{table*}

The number of edges between subgraphs, $s_b$ in Table~\ref{tab:pathfinderAndes}, did not show any real impact on the accuracy of PEF-CCDr. This is because when we fuse DAGs from clusters we also correct their structures learned in the E-step. Therefore, even if we cut some edges in the P-step, which may alter the subDAG structures, we can still correct them in the F-step. Therefore, our PEF-CCDr method has some tolerance for errors in the first two steps. Even if the full DAG does not have a clear cluster structure, in which case many edges will be cut in the P-step, PEF-CCDr can still recover a reasonable amount of these edges. This is demonstrated by the comparable performance of our method on DAGs with a different $s_b$ in the table.

On the other hand, performance of PEF-CCDr clearly depends on the structure learning algorithm plugged in the E-step. In the final F-step, we only remove or flip within-subgraph edges, so the missing edges within any subgraph introduced in the E-step will never be added back in the fusion step. In addition, the learned subgraph structure may also affect our choice for the candidate set $A$ (Section~\ref{sec:findA}) and thus the final accuracy. 

\subsubsection{Recovery rate of the fusion step}\label{sec:fusionPerformance}

To examine the role of the fusion step, we compare DAGs learned by our full PEF method with DAGs learned from the first two steps only, i.e. the partition step and the estimation step. We call the latter PE-CCDr. Table~\ref{tab:pathfinderAndesWithin} reports the comparison between these two methods. The rows for PEF-CCDr report the percentage of change in each accuracy metric relative to PE-CCDr. Using ANDES$(5,0.1)$ with $s_b=169$ as an example, PEF-CCDr predicted 63\% more expected edges with the SHD 55\% smaller than that of PE-CCDr. It is clear from the results that the fusion step always improved the structure of an estimated DAG with increased E, JI and decreased R, FP, SHD. 

\begin{table*}[!htb]
\begin{center}
\caption{Accuracy comparison between PE-CCDr and PEF-CCDr}
\label{tab:pathfinderAndesWithin}       
{\renewcommand{\arraystretch}{0.95}
\begin{tabular}{llrrrrrr}
\noalign{\smallskip}\hline\noalign{\smallskip}
 $(s_0, s_b)$ & Method & P & E & R & FP &  SHD & JI \\
\noalign{\smallskip}\hline\noalign{\smallskip}
\multicolumn{8}{c}{PATHFINDER$(5)$, $p = 545$} \\
\noalign{\smallskip}
{$(975, 0)$} & PE-CCDr & 818.4 & 250.6 & 146.4 & 421.4 & 1145.8 & 0.163 \\
& PEF-CCDr($\%$) & $-19$ & 2 & $-16$ & $-33$ & $-12$ & 14 \\
{$(1073, 98)$} & PE-CCDr & 830.0 & 275.0 & 134.3 & 420.7 & 1218.7 & 0.169 \\
& PEF-CCDr($\%$) & $-7$ & 31 & $-11$ & $-31$ & $-18$ & 45 \\

\noalign{\smallskip}
\multicolumn{8}{c}{ANDES$(5)$, $p = 1115$} \\
\noalign{\smallskip}
{$(1690, 0)$} & PE-CCDr & 1652.0 & 873.3 & 456.4 & 322.3 & 1139.0 & 0.354 \\
& PEF-CCDr($\%$) & $-5$ & 36 & $-51$ & $-53$ & $-43$ & 63 \\
{$(1859, 169)$} & PE-CCDr & 1691.8 & 860.5 & 452.3 & 379.0 & 1377.5 & 0.320 \\
& PEF-CCDr($\%$) & 4 & 63 & $-59$ & $-54$ & $-55$ & 98 \\

\noalign{\smallskip}
\multicolumn{8}{c}{DIABETES$(5)$, $p = 2065$} \\
\noalign{\smallskip}
{$(3010, 0)$} & PE-CCDr & 3119.5 & 1281.3 & 1050.6 & 787.6 & 2516.3 & 0.264 \\
& PEF-CCDr($\%$) & $-11$ & 23 & $-26$ & $-47$ & $-26$ & 42 \\
{$(3311, 301)$} & PE-CCDr & 3058.5 & 1256.6 & 994.6 & 807.3 & 2861.7 & 0.246 \\
& PEF-CCDr($\%$) & 5 & 60 & $-29$ & $-39$ & $-37$ & 82 \\

\noalign{\smallskip}
\multicolumn{8}{c}{PIGS$(5)$, $p = 2205$} \\
\noalign{\smallskip}
{$(2960, 0)$} & PE-CCDr & 3290.1 & 1632.7 & 834.7 & 822.7 & 2150.0 & 0.354 \\
& PEF-CCDr($\%$) & $-15$ & 18 & $-35$ & $-59$ & $-37$ & 42 \\
{$(3256, 296)$} & PE-CCDr & 3312.9 & 1574.7 & 825.4 & 912.8 & 2594.1 & 0.315 \\
& PEF-CCDr($\%$) & $-4$ & 47 & $-41$ & $-58$ & $-49$ & 77 \\

\noalign{\smallskip}
\multicolumn{8}{c}{LINK$(5)$, $p = 3620$} \\
\noalign{\smallskip}
{$(5625, 0)$} & PE-CCDr & 5432.8 & 2514.9 & 1432.3 & 1485.6 & 4595.7 & 0.294 \\
& PEF-CCDr($\%$) & $-8$ & 28 & $-32$ & $-44$ & $-29$ & 47 \\
{$(6188, 563)$} & PE-CCDr & 5471.0 & 2422.9 & 1459.3 & 1588.8 & 5353.9 & 0.262 \\
& PEF-CCDr($\%$) & 7 & 66 & $-40$ & $-40$ & $-42$ & 91 \\

\noalign{\smallskip}
\multicolumn{8}{c}{Mix$(5)$, $p = 1910$} \\
\noalign{\smallskip}
{$(2852, 0)$} & PE-CCDr & 2848.1 & 1330.9 & 765.6 & 751.6 & 2272.7 & 0.305 \\
& PEF-CCDr($\%$) & $-8$ & 27 & $-31$ & $-46$ & $-31$ & 46 \\
{$(3138, 286)$} & PE-CCDr & 2880.4 & 1313.1 & 779.1 & 788.2 & 2613.1 & 0.279 \\
& PEF-CCDr($\%$) & 3 & 53 & $-36$ & $-41$ & $-39$ & 75 \\
\noalign{\smallskip}\hline
\end{tabular}%
}
\end{center}
\end{table*}

The results in Table~\ref{tab:pathfinderAndesWithin} show that as $s_b$ increased, the fusion step recovered an increasing number of expected edges. The number of expected edges recovered by the fusion step can reach $60\%$ of that recovered in the first two steps, such as for ANDES$(5,0.1)$, DIABETES$(5,0.1)$ and LINK$(5, 0.1)$.
In addition, we see that our fusion step not only recovered expected edges, but also was able to remove reversed and false positive edges. 
Across different cases, the F-step reduced $30\%$ to $60\%$ FPs and $10\%$ to $60\%$ Rs, which substantially improved the structure learning accuracy.

All these observed improvements in accuracy demonstrate the critical role of the fusion step. Not only does it add edges cut by the P-step back to the full DAG, but also gets rid of false positive edges produced by the E-step.  
This suggests that the fusion step can largely correct the mistakes made by the first two steps, and thus our PEF method may handle networks with a moderate number of between-subgraph edges, relaxing the assumption on a block structure of the true DAG to some degree.

\subsection{Comparison with the PC algorithm}\label{sec:PEFpcalgSim}

In this section, we test our PEF framework with the PC algorithm used for the E-step, which we call PEF-PC. The PC algorithm is a well-known constraint-based method that outputs a PDAG in general. This will complement our comparison with CCDr in the previous subsection, which estimates a DAG via a score-based approach.

In our experiments, we used the PC algorithm in the \texttt{pcalg} package \citep{kalisch2012causal}. An important tuning parameter of PC is the significance level $\alpha$ for conditional independence tests, which controls the sparsity of an estimated graph: The smaller the $\alpha$, the sparser the estimated graph and the faster the algorithm. With the default setting $\alpha = 0.05$, PC took too long, more than 24 hours, to learn some DAGs like the PATHFINDER(5) networks. Furthermore, for high-dimensional data, a big $\alpha$ usually results in too many false positive edges in the graph learned by the PC algorithm. In order to make an informative comparison, we set $\alpha=10^{-4}$ so that the PC algorithm can produce quite accurate PDAGs within a reasonable amount of time. 
Another tuning parameter is the maximal size (\texttt{m.max}) of the conditioning sets that are considered in a conditional independence test. The default value of this parameter is infinity, but with this default value, it took up to 6 hours to run PC on a single data set. Thus, in our experiment, we limited this value to 3. We also tried increasing \texttt{m.max} to 5, and got similar results with slightly lower accuracy but much longer running time. The same data for the comparisons in Tables~\ref{tab:timePefCcdr} and \ref{tab:pathfinderAndes} were used in this experiment as well. The parameter choices for PC in our E-step were the same as those for running PC on full data.


\begin{table*}[t]
\begin{center}
\caption{Timing comparison (in minutes) between PC and PEF-PC}
\label{tab:pcalgTiming}       
{\renewcommand{\arraystretch}{0.95}
\begin{tabular}{ll | r | rrrr | r }
\noalign{\smallskip}\hline\noalign{\smallskip}
& & PC & \multicolumn{4}{  c|}{PEF-PC} & \\
\noalign{\smallskip}\cline{3-8}\noalign{\smallskip}
Network & $p$ &  T  & T & P & E & F & $r_T$ \\
\noalign{\smallskip}\hline\noalign{\smallskip}
\noalign{\smallskip}
PATHFINDER(5, 0) & 545 & 3.50 & 1.89 & 0.01 & 1.85 & 0.03 & 1.85 \\
PATHFINDER(5, 0.1) & 545 & 3.54 & 1.64 & 0.01 & 1.54 & 0.09 & 2.16 \\
\noalign{\smallskip}
\noalign{\smallskip}
ANDES(5, 0) & 1115 & 0.52 & 0.32 & 0.01 & 0.04 & 0.27 & 1.63 \\
ANDES(5, 0.1) & 1115 & 0.59 & 0.39 & 0.02 & 0.04 & 0.33 & 1.51 \\
\noalign{\smallskip}
\noalign{\smallskip}
DIABETES(5, 0) & 2065 & 2.67 & 0.57 & 0.06 & 0.22 & 0.29 & 4.68 \\
DIABETES(5, 0.1) & 2065 & 2.82 & 0.67 & 0.05 & 0.18 & 0.44 & 4.21 \\
\noalign{\smallskip}
\noalign{\smallskip}
PIGS(5, 0) & 2205 & 4.33 & 1.01 & 0.07 & 0.62 & 0.32 & 4.29 \\
PIGS(5, 0.1) & 2205 & 4.87 & 0.96 & 0.07 & 0.51 & 0.38 & 5.07 \\
\noalign{\smallskip}
\noalign{\smallskip}
LINK(5, 0) & 3620 & 8.37 & 1.12 & 0.16 & 0.35 & 0.61 & 7.47 \\
LINK(5, 0.1) & 3620 & 9.00 & 1.36 & 0.16 & 0.36 & 0.84 & 6.62 \\
\noalign{\smallskip}
\noalign{\smallskip}
MIX(5, 0) & 1910 & 3.05 & 1.82 & 0.06 & 0.93 & 0.83 & 1.68 \\
MIX(5, 0.1) & 1910 & 3.70 & 1.75 & 0.06 & 1.08 & 0.61 & 2.11 \\
\noalign{\smallskip}\hline\noalign{\smallskip}
\end{tabular}%
}
\\ ({See Table~\ref{tab:timePefCcdr} for the definitions of T, P, E, F, and $r_T$.})
\end{center}
\end{table*}

Table~\ref{tab:pcalgTiming} compares the running time between PC and PEF-PC with paralleling the E-step. As described above, we did fine tuning on the parameters of PC to improve its speed, and consequently, the algorithm ran very fast on these data sets. Even though, PEF-PC was usually 2 to 8 times faster. 
The running time improvement here was not as substantial as that for the CCDr algorithm, probably because of the different ways these two algorithms scale with the graph size. 

Next, we compared the estimation accuracy between PC and PEF-PC. To confirm the effect of the fusion step in our method, we also compared with PE-PC, which only includes the first two steps of our PEF framework. Table~\ref{tab:pcalgAccuracy} reports the detailed accuracy metrics. Similar to the results in the comparison with CCDr, we observe significant improvement in accuracy of PEF-PC over PC. For all the networks tested, the Jaccard index of PEF-PC was much higher and the SHD of PEF-PC was much lower than PC. 
Consistent with Table~\ref{tab:pathfinderAndesWithin}, the fusion step of PEF-PC substantially improved the results from the P-step and the E-step, by recovering more expected edges and correcting many reversed edges. Take the ANDES$(5)$ network with $s_b=0$ as an example. Our PEF method found $30\%$ more expected edges, while reducing reversed edges by more than $80\%$, compared to the other two competitors.

\begin{table*}[!htb]
\caption{Accuracy comparison between PC, PE-PC and PEF-PC}
\label{tab:pcalgAccuracy}       
\centering
{\renewcommand{\arraystretch}{0.8}
\begin{tabular}{llrrrrrr}
\noalign{\smallskip}\hline\noalign{\smallskip}
$(s_0, s_b)$ & Method & P & E & R & FP &  SHD & JI \\
\noalign{\smallskip}\hline\noalign{\smallskip}
\multicolumn{8}{c}{PATHFINDER$(5)$, $p = 545$} \\
\noalign{\smallskip}
{$(975, 0)$} & PC & 438.0 & 154.6 & 190.1 & 93.3 & 913.7 & 0.123 \\
& PE-PC & 436.9 & 154.6 & 189.4 & 92.9 & 913.3 & 0.123 \\
& PEF-PC & 434.9 & 255.8 & 85.0 & 94.1 & 813.3 & 0.222 \\
{$(1073, 98)$} & PC & 521.9 & 205.1 & 218.4 & 98.4 & 966.3 & 0.148 \\
& PE-PC & 492.0 &176.1 & 192.5 & 123.4 & 1020.3 & 0.127 \\
& PEF-PC & 660.8 & 342.8 & 99.8 & 218.2 & 948.4 & 0.247 \\

\noalign{\smallskip}
\multicolumn{8}{c}{ANDES$(5)$, $p = 1115$} \\
\noalign{\smallskip}
{$(1690, 0)$} &  PC & 1483.0 & 1143.8 & 318.3 & 20.9 & 567.1 & 0.564  \\
& PE-PC & 1398.4 & 1050.6 & 307.1 & 40.7 & 680.1 & 0.516 \\
& PEF-PC & 1520.7 & 1423.1 & 49.2 & 48.4 & 315.3 & 0.796 \\
{$(1859, 169)$} & PC & 1635.4 & 1230.7 & 383.1 & 21.6 & 649.9 & 0.544 \\
& PE-PC & 1387.8 & 993.7 & 333.5 & 60.6 & 925.9 & 0.441 \\
& PEF-PC & 1714.3 & 1589.1 & 48.0 & 77.2 & 347.1 & 0.801 \\

\noalign{\smallskip}
\multicolumn{8}{c}{DIABETES$(5)$, $p = 2065$} \\
\noalign{\smallskip}
{$(3010, 0)$} & PC & 2563.0 &1875.9 & 669.6 & 17.5 & 1151.6 & 0.507 \\
& PE-PC & 2506.2 & 1807.5 & 653.5 & 45.2 & 1247.7 & 0.487 \\
& PEF-PC & 2601.1 & 2192.0 & 353.1 & 56.0 & 874.0 & 0.641 \\
{$(3311, 301)$} & PC & 2850.5 & 2071.7 & 750.2 & 28.6 & 1267.9 & 0.507 \\
& PE-PC & 2466.4 & 1676.9 & 679.3 & 110.2 & 1744.3 & 0.409 \\
& PEF-PC & 3009.5 & 2533.1 & 324.4 & 152.0 & 929.9 & 0.669 \\

\noalign{\smallskip}
\multicolumn{8}{c}{PIGS$(5)$, $p = 2205$} \\
\noalign{\smallskip}
{$(2960, 0)$} & PC & 2556.6 & 1881.6 & 638.5 & 36.5 & 1114.9 & 0.519 \\
& PE-PC & 2497.2 & 1797.9 & 649.9 & 49.4 & 1211.5 & 0.492 \\
& PEF-PC & 2586.6 & 2256.0 & 263.2 & 67.4 & 771.4 & 0.686 \\
{$(3256, 296)$} & PC & 2859.6 & 2116.5 & 707.5 & 35.6 & 1175.1 & 0.530 \\
& PE-PC & 2525.8 & 1696.3 & 695.7 & 133.8 & 1693.5 & 0.415 \\
& PEF-PC & 2989.4 & 2613.3 & 228.3 & 147.8 & 790.5 & 0.720 \\

\noalign{\smallskip}
\multicolumn{8}{c}{LINK$(5)$, $p = 3620$} \\
\noalign{\smallskip}
{$(5625, 0)$} & PC & 4752.8 & 3480.3 & 1115.8 & 156.7 & 2301.4 & 0.505 \\
& PE-PC & 4510.0 & 3182.4 & 1098.7 & 228.9 & 2671.5 & 0.458 \\
& PEF-PC & 4734.4 & 4372.8 & 218.9 & 142.7 & 1394.9 & 0.730 \\
{$(6188, 563)$} & PC & 5244.1 & 3818.8 & 1304.5 & 120.8 & 2490.0 & 0.502 \\
& PE-PC & 4361.3 & 2861.3 & 1199.1 & 300.9 & 3627.6 & 0.372 \\
& PEF-PC & 5434.0 & 4924.0 & 228.4 & 281.6 & 1545.6 & 0.735 \\

\noalign{\smallskip}
\multicolumn{8}{c}{Mix$(5)$, $p = 1910$} \\
\noalign{\smallskip}
{$(2852, 0)$} & PC & 2376.6 & 1709.4 & 606.7 & 60.5 & 1203.1 & 0.486 \\
& PE-PC & 2251.5 & 1564.3 & 588.6 & 98.6 & 1386.3 & 0.442 \\
& PEF-PC & 2409.1 & 2116.7 & 201.3 & 91.1 & 826.4 & 0.673 \\
{$(3138, 286)$} & PC & 2621.5 & 1875.5 & 679.2 & 66.8 & 1329.3 & 0.483 \\
& PE-PC & 2280.3 & 1514.4 & 617.0 & 148.9 & 1772.5 & 0.388 \\
& PEF-PC & 2766.3 & 2380.2 & 202.8 & 183.3 & 941.1 & 0.676 \\
\noalign{\smallskip}\hline
\end{tabular}%
}
\end{table*}

\begin{remark}
Comparing the results in this section with those in Section~\ref{sec:PEFccdrSim}, it appears that the PEF-PC method outperformed the PEF-CCDr method in terms of accuracy for most of the networks except PATHFINDER(5). This is because the PC algorithm had higher accuracy than the CCDr algorithm on these data. Such differences match our expectation that performance of the PEF framework will depend on the algorithm used in the E-step.
On the other hand, our PEF framework showed substantial advantages over both algorithms, demonstrating the robustness of our divide-and-conquer strategy regardless of the performance of the DAG learning algorithm used in the E-step.
\end{remark}

\section{Discussion}\label{sec:PEFdiscussion}

We have developed a divide-and-conquer framework for structure learning of massive Bayesian networks from continuous data.
The key novel step in our method is the fusion step, which merges the subgraphs learned from subsets of nodes partitioned by a modified clustering algorithm. Our numerical results suggest that this fusion step can correct and fix the DAG structure damaged by the partition step, so that the overall accuracy of the PEF method is seen to be much higher than the structure learning algorithm used in the estimation step. We also observed quite significant boost in speed, ranging from a few folds to orders-of-magnitude.

There are certain limitations of our current design and implementation of the PEF method. First of all, in the partition step, we need to calculate and store the dissimilarity matrix for all pairs of nodes. When the number of nodes $p$ is really large, this becomes memory-intensive. A promising potential solution to this issue is to borrow ideas from the subsample clustering method for big data \citep{marchetti2016iterative}. We may subsample a small fraction of the nodes for clustering, and then assign the remaining large number of nodes based on the clustering of the subsample, which can be implemented in a sequential way. For the fusion step, our current implementation takes as input the correlation matrix of the data columns. Again, when the number of nodes is too big, we may implement the algorithm to calculate correlations whenever needed, instead of pre-computing all correlations. Our current fusion step was implemented with the \texttt{Rcpp} package \texttt{Armadillo}. If we code it in pure C++, the speed of the fusion step may be further improved.

At a conceptual level, it seems straightforward to generalize the PEF method to discrete Bayesian networks. For discrete data, one can still use our clustering method, with a suitable similarity measure, for the partition step, and plug in an appropriate structure learning algorithm in the estimation step. As for the fusion step, the conditional independence test is no longer for zero partial correlations, instead we may use the $G^2$ test for discrete data as in the PC algorithm. Finally we may substitute linear regression with the multinomial logistic regression as used in \cite{gu2017penalized} for BIC-based edge selection (Section~\ref{sec:highDimBIC}). This is left as future work.

\appendix
\section{Appendix}

\subsection{Partial correlation} \label{sec:partialCorr}

The partial correlation between $X$ and $Y$ given $\boldsymbol{Z}$, $\rho_{XY\cdot\boldsymbol{Z}}$,  can be calculated using their covariance matrix. Let $k$ be the size of $\boldsymbol{Z}$,
$\Sigma$ be the covariance matrix of $(X, Y, \boldsymbol{Z})$, and $\Omega = (\omega_{ij})_{(k+2)\times (k+2)} = \Sigma^{-1}$ be the precision matrix.  Then the partial correlation 
\begin{equation*}\label{eq:partialCorrMatrix}
\rho_{XY\cdot \boldsymbol{Z}} = -\frac{\omega_{12}}{\sqrt{\omega_{11}\omega_{22}}},
\end{equation*}
and for Gaussian random variables, 
\begin{equation*}\label{eq:CI-partialCorr}
\mathcal{I}_P(X; Y|\boldsymbol{Z}) \qquad\Longleftrightarrow\qquad  \rho_{XY\cdot \boldsymbol{Z}} = 0.
\end{equation*}
In order to test the hypothesis $H_0: \rho_{XY\cdot \boldsymbol{Z}}=0$, we apply the Fisher z-transformation,
\begin{equation*}\label{eq:zTransform}
z(X, Y | \boldsymbol{Z}) = \frac{1}{2}\log\left(\frac{1+\hat{\rho}_{XY\cdot \boldsymbol{Z}}}{1-\hat{\rho}_{XY\cdot \boldsymbol{Z}}}\right),
\end{equation*}
where $\hat{\rho}_{XY\cdot \boldsymbol{Z}}$ is the estimated partial correlation calculated from sample covariance matrix of $(X,Y,\boldsymbol{Z})$.
Given a significance level $\alpha$, we reject the null hypothesis $H_0$ if
\begin{equation*}\label{eq:zTestForPartialCorr}
\sqrt{n-k-3}\lvert z(X, Y | \boldsymbol{Z})\rvert > \Phi^{-1}\left(1-\alpha/2\right),
\end{equation*}
where $n$ is the number of observations and $\Phi$ is the cdf of $\mathcal{N}(0, 1)$.

\subsection{Proof of Proposition~\ref{thm:residual}} \label{sec:ProofPropo}

By properties of a joint Gaussian distribution, we can write
\begin{align}\label{eq:xi}\tag{A.1}
X_i=\sum_{k\in A}\widetilde{\beta}_{ki} X_k + R_{i\cdot A},
\end{align}
where $R_{i\cdot A} \perp X_A$ (independence). Similarly, regressing $X_j$ onto $X_{A\cup B\cup \{i\}}$, we arrive at
\begin{align}\label{eq:xj}\tag{A.2}
X_j=\sum_{k\in A\cup B}\widetilde{\beta}_{kj} X_k + \widetilde{\beta}_{ij} X_i+\widetilde{\veps}_j,
\end{align} 
with $\widetilde{\veps}_j \perp X_{A\cup B\cup \{i\}}$ and thus $\widetilde{\veps}_j\perp R_{i\cdot A}$. Plugging \eqref{eq:xi} into \eqref{eq:xj} to eliminate $X_i$, we have
\begin{align}\tag{A.3}
X_j=\sum_{k\in A\cup B}\widetilde{\gamma}_{kj} X_k + \widetilde{\beta}_{ij} R_{i\cdot A}+\widetilde{\veps}_j,
\end{align} 
for some $\widetilde{\gamma}_{kj}$'s after rearranging terms in the summation. Denote by $R_{i\cdot A\cdot B}$ the residual of regressing $R_{i\cdot A}$ onto $X_B$. Since $R_{i\cdot A}\perp X_A \mid X_B$ by assumption, the coefficient 
\begin{align*}
\widetilde{\beta}_{ij}=\frac{\E(R_{j\cdot B} R_{i\cdot A\cdot B})}{\E(R_{i\cdot A\cdot B})^2}=\frac{\E(R_{j\cdot B} R_{i\cdot A})}{\E(R_{i\cdot A\cdot B})^2}=\frac{\text{cov}(R_{j\cdot B}, R_{i\cdot A})}{\text{var}(R_{i\cdot A\cdot B})},
\end{align*}
where the second equality is due to $R_{j\cdot B} \perp \E(R_{i\cdot A}\mid X_B)$.
By Theorem~\ref{thm:CIstruct}, $\widetilde{\beta}_{ij}\ne 0$, because otherwise 
$\mathcal{I}_P(X_j; X_i | X_{A\cup B})$, and thus $\text{cov}(R_{j\cdot B},R_{i\cdot A})\ne 0$. The proof is complete.

\subsection{RIC for model selection} \label{sec:RIC}

Recall we want to compare three models, $M_0,M_1,M_2$, defined in \eqref{eq:modelCompare}.
Suppose the current DAG is $\calG$, which has no edge between $i$ and $j$. Now consider the following two linear models
\begin{align}
X_i =& \beta_{ji}X_j + \sum_{k\in \Pi_i^{\mathcal{G}}}\beta_{ki}X_k + \veps_i,  \label{eq:lmJ2i}\tag{A.4} \\
X_j =& \beta_{ij}X_i + \sum_{k\in \Pi_j^{\mathcal{G}}}\beta_{kj}X_k + \veps_j. \label{eq:lmI2j}\tag{A.5}
\end{align}
Then, $M_0$ is equivalent to $\beta_{ij} = \beta_{ji} = 0$, 
$M_1$ equivalent to $\beta_{ij} \neq 0$ and $\beta_{ji} = 0$, and
$M_2$ equivalent to $\beta_{ij} = 0$ and $\beta_{ji} \neq 0$.
Note that when undirected edges exist, we consider all neighbors as the parents.

In order to choose from the three models, we calculate their RIC scores. 
In our implementation, we find least-squares estimates (LSEs) of the regression coefficients for \eqref{eq:lmJ2i} and \eqref{eq:lmI2j}. Let $\ell_{ji}$ be the log-likelihood evaluated at the LSE under the linear model \eqref{eq:lmJ2i}, and $\ell_{0i}$ the log-likelihood under \eqref{eq:lmJ2i} when $\beta_{ji} = 0$.  Similarly, $\ell_{ij}$ denotes the log-likelihood at the LSE for the linear model \eqref{eq:lmI2j}, and $\ell_{0j}$ the log-likelihood when $\beta_{ij} = 0$. Since the structure of $\calG$ is identical except for the node pair $(i, j)$, these four likelihood scores are sufficient for comparing $M_0$, $M_1$ and $M_2$. Let $\ell(M_i)$ be the log-likelihood of $M_i$ for $i=0,1,2$. Then we have $\ell(M_0)=\ell_{0i}+\ell_{0j}$, $\ell(M_1)=\ell_{0i}+\ell_{ij}$, and  $\ell(M_2)=\ell_{ji}+\ell_{0j}$. Thus, the RIC selection criterion \eqref{eq:BICcriterion} is equivalent to $2\min\{\ell_{ij} - \ell_{0j}, \ell_{ji} - \ell_{0i} \}> \lambda$, where $\lambda$ is the penalty parameter in \eqref{eq:BIC}.
The motivation for this criterion is to add an edge between $i$ and $j$ only when $i\not\perp j | \Pi_i^{\mathcal{G}}$ and  $i\not\perp j | \Pi_j^{\mathcal{G}}$ (Theorem~\ref{thm:CIstruct}).

\bibliographystyle{asa}
\bibliography{references}

\end{document}

%% file: NewCommands.tex
\newcommand{\bfx}{\mathbf{x}}

\newcommand{\calC}{\mathcal{C}}

\newcommand{\calG}{\mathcal{G}}

\newcommand{\R}{\mathbb{R}}

\newcommand{\veps}{\varepsilon}

\newcommand{\Omg}{\Omega}

\newcommand{\E}{\mathbb{E}}

\newcommand{\defi}{\mathop{:=}} 
  
\newcommand{\dnorm}{\mathcal{N}}

\newcommand{\trans}{\mathsf{T}}

\DeclareMathOperator*{\argmin}{argmin}
\DeclareMathOperator*{\argmax}{argmax}

\DeclareMathOperator{\diag}{diag}

\DeclareMathOperator{\cor}{cor}

\RequirePackage{tikz}